\documentclass{article}
\usepackage[utf8]{inputenc}
\usepackage{stywhispers}
\usepackage{amsmath}
\usepackage{epsfig}
\usepackage{amssymb}
\usepackage{booktabs}
\usepackage{multirow}
\usepackage{graphicx}
\usepackage{rotating}
\usepackage{subcaption}
\usepackage{stfloats}
\usepackage{float}
\usepackage{graphicx}
\usepackage{url}
\usepackage{xcolor}
\usepackage[table]{xcolor}
\usepackage{graphicx}
\usepackage{multirow}
\usepackage{array}
\usepackage[table]{xcolor}
\usepackage{pgf}
\usepackage{subcaption}
\usepackage{caption}

\definecolor{heatred}{RGB}{178,24,43}
\definecolor{heatgreen}{RGB}{79, 148, 84} 

\newcommand{\heatcell}[2]{%
  \pgfmathtruncatemacro{\heatpct}{round(min(55,55*abs(#1)/3))}%
  \ifdim #1pt<0pt
    \edef\heatshade{heatred!\heatpct!white}%
    \expandafter\cellcolor\expandafter{\heatshade}#2%
  \else
    \ifdim #1pt>0pt
      \edef\heatshade{heatgreen!\heatpct!white}%
      \expandafter\cellcolor\expandafter{\heatshade}#2%
    \else
      \cellcolor{white}#2%
    \fi
  \fi
}

\newcommand{\spm}{\scalebox{0.7}{$\pm$}}

\title{Band-Selection Stability and Semantic Segmentation Performance: A Study on Hyperspectral City}

\name{Jiarong Li~\textsuperscript{1,†},
Imad Ali Shah~\textsuperscript{1,†},
Enda Ward~\textsuperscript{2},
Martin Glavin~\textsuperscript{1},
Edward Jones~\textsuperscript{1} and
Brian Deegan~\textsuperscript{1}
\thanks{This work was supported, in part, by Taighde Éireann -- Research Ireland grants 13/RC/2094\_P2 and 18/SP/5942, and Valeo Vision Systems.
\newline † Equal Contribution and correspondence (i.shah2@universityofgalway.ie)}
}
\address{\textsuperscript{1} School of Engineering and Ryan Institute, University of Galway, Ireland\\
\textsuperscript{2} Valeo Vision Systems, Tuam, Ireland}

\begin{document}
\maketitle

\begin{abstract}
Resource constraints make high-dimensional hyperspectral imaging challenging in autonomous perception, motivating the use of band selection methods. However, the sensitivity of band-selection methods to sampled data and their relationship to semantic segmentation models (SSMs) remain underexplored. This study evaluates six band~selection methods on ten independently sampled, class-balanced region-of-interest (ROI) sets, yielding 60 top-25 band subsets from the Hyperspectral City V2 (128 bands:~450-950nm) dataset. Top-$K$ bands ($K\in\{3,5, ... 13\}$) from the first three ROI sets are evaluated with three SSMs against the corresponding 128-band baseline. Experiments show that intra-method stability is method-dependent: Sim-LP shows the highest stability (pairwise Jaccard similarity) and, together with JMIM+CSNR, yields the best segmentation results.~Top-$K$~based~SSMs remain competitive with baselines, with gains of up to 2.01 mIoU and 1.72 mF1 points, and 18--22x faster CPU inference for~$K=9$.~However, performance does not improve monotonically with $K$, and stability shows no consistent association with SSM performance. These findings suggest that intra-method stability is informative but an unreliable indicator of downstream segmentation performance, highlighting the need to evaluate band-selection methods across repeated samples, subset sizes, and SSMs.
\end{abstract}


\vspace{-1.5ex}
\section{Introduction}
\vspace{-1.5ex}
\label{sec:intro}

Hyperspectral imaging (HSI) captures dense spectral information that can provide material information beyond conventional RGB imaging. This capability is relevant to semantic segmentation in urban-driving scenes, where different materials may exhibit similar visible appearance~\cite{shah2025hyperspectral}. However, HSI is high-dimensional, computationally demanding, and often contains spectral redundancy~\cite{yuan2014hyperspectral}. Band selection is therefore used to retain a small but informative set of wavelengths.

Band-selection criteria are typically estimated from sampled pixels or an image region of interest (ROI), and depend on the class distribution represented in the sample. This issue is particularly important in automotive segmentation datasets with severe class imbalance~\cite{shah2024hyperspectral}.~In the Hyperspectral City~V2 (H-City)~\cite{shen2024urban} dataset used in this study, Road accounts for 22.43\% of labeled pixels versus 0.02\% for Train. Band-selection criteria estimated directly on such data may therefore bias selection toward frequent classes. Moreover, a stable method that reliably selects the same bands when the sampled regions change does not necessarily imply that those bands are optimal for segmentation performance. Conversely, different subsets may provide similar performance due to inter-band spectral redundancy. This study therefore addresses two questions: how strongly do common band-selection methods vary under repeated class-balanced ROI sampling, and does greater intra-method stability correspond to better semantic segmentation models (SSMs) performance?

Six band-selection methods are applied to ten class-balanced ROI sets, producing 60 top-25 subsets from the H-City dataset. Top-$K$ bands, where $K\in\{3,5, ... 13\}$, from three~ROI sets are then evaluated with three SSMs against 128-band baselines.~A frequency-consensus ranking derived from comparatively stable methods is included as an exploratory aggregate baseline.  The study consequently contributes: (1) a repeated-sampling analysis of band-selection stability and spectral recurrence; (2) evaluation across $K$-subsets and three SSMs; and (3) analysis of the relationship between intra-method stability and downstream SSM performance, including the accuracy-inference-time trade-off.

\vspace{-1.5ex}
\section{Related Work}
\vspace{-1.5ex}
\label{sec:format}
\subsection{HSI and Band-Selection in Autonomous Driving}
\vspace{-1ex}
HSI is increasingly investigated as a source of complementary spectral information for urban-scene understanding~\cite{shah2025hyperspectral}. Huang et al.~\cite{huang2020weakly} used hyperspectral information to refine coarse urban-scene annotations under weak supervision. Subsequent works evaluated different baseline SSM performance~\cite{shah2024hyperspectral} on the four publicly available annotated automotive HSI datasets considered in that study.

Band-selection has also been investigated for vulnerable-road-user perception~\cite{li2026csnr} and compared with RGB~\cite{li2025hyperspectral}. In contrast to independently applied filter methods, embedded approaches learn spectral gates or selectors jointly with the downstream model~\cite{zimmer2024embedded, shah2026learnable}. Embedded methods can optimize bands for a particular architecture and objective, but the learned selection may also depend on model initialization and optimization. They are therefore related to, but distinct from, the sampling-stability problem considered in this work.

\vspace{-1.5ex}
\subsection{Hyperspectral Band-Selection}
\vspace{-1ex}

Band-selection methods differ in class supervision and handling of inter-band redundancy.~Minimum Redundancy Maximum Relevance (mRMR)~\cite{peng2005feature} balances class relevance and inter-band redundancy, whereas Conditional Mutual Information Maximization (CMIM)~\cite{fleuret2004fast} and Joint Mutual Information Maximization (JMIM)~\cite{bennasar2015feature} rank candidates using conditional and joint mutual-information criteria, respectively. These supervised criteria can represent nonlinear statistical dependence, but their rankings depend on the labeled sample and mutual information estimation. Similarity-based Linear Prediction (Sim-LP)~\cite{du2008similarity} is an unsupervised alternative that favors bands poorly reconstructed from selected bands, discouraging redundancy. Li et al.~\cite{li2026csnr} combined JMIM with patch-based contrast signal-to-noise ratio (CSNR) and correlation-based duplicate control. JMIM+CSNR combines JMIM and CSNR, while JMIM+CSNR+Corr additionally applies Pearson-correlation-based duplicate removal.

\vspace{-1.5ex}
\subsection{Band-Selection Stability vs SSM Performance}
\vspace{-1ex}

Band-selection stability measures the degree of agreement among the selected variables when the data sample is perturbed~\cite{nogueira2018stability}. It is commonly estimated by repeating the selection method and comparing the resulting subsets: Pairwise Jaccard similarity~\cite{jaccard1912distribution} provides a direct overlap measure for fixed-size subsets. Liu et al.~\cite{liu2017stability} constructed hyperspectral perturbation datasets with controlled sample overlap and examined band-selection stability alongside classification accuracy. However, the relationship between intra-method stability under repeated class-balanced ROI sampling and downstream SSM performance remains underexplored, especially for the automotive HSI domain. 

\vspace{-1.5ex}
\section{Experimental Protocol}
\vspace{-1ex}
\label{sec:pagestyle}

\subsection{Dataset and Class-Balanced ROI Sampling}
\vspace{-1ex}
The experiments used H-City, whose 128-band, 450-950nm range and pixel-level annotations for 19 urban classes provide a comparatively large spectral search space for intra-method band-selection stability analysis. Due to the different annotation granularity level in the original H-City's validation and test sets, ROI sampling used the training split: SSMs were trained on the training and evaluated on the validation split. The class-balanced ROI-sampling procedure was repeated using ten different sampling seeds. Each sampling run produced one ROI subset, sampled under the same class-balancing rule.

\vspace{-1.5ex}
\subsection{Intra-Method Stability Characterization}
\vspace{-1ex}

Six band-selection methods were applied independently to each ROI set: JMIM, JMIM+CSNR, JMIM+CSNR+Corr, CMIM, mRMR, and Sim-LP. For method $m$ and ROI set $r$, the top-$L$ ($L=25$) bands in the resulting ranking formed the subset $S_{m,r}^{L}$. Applying the six methods to the ten ROI sets produced 60 top-$L$ subsets. Band $b$'s recurrence under method $m$ was defined as $R_m(b) = \sum_{r=1}^{10}\mathcal{I}\left(b\in S_{m,r}^{L}\right)$, where $\mathcal{I}(\cdot)$ is the indicator function, characterizing how consistently each band was selected by $m$ across the ten ROI sets. Intra-method stability was quantified using pairwise Jaccard similarity among the ten top-$L$ subsets produced by each method $m$. For ROI sets $r_i$ and $r_j$:

\vspace{-1ex}
$$
J_m(r_i,r_j)
=
\frac{
\left|S_{m,r_i}^{25}\cap S_{m,r_j}^{25}\right|
}{
\left|S_{m,r_i}^{25}\cup S_{m,r_j}^{25}\right|
}
$$

Each method yield $\binom{10}{2}=45$ pairwise Jaccard similarities. The median and interquartile range summarized the central value and spread of this distribution. For reference, the expected Jaccard similarity of two random 25-of-128 subsets is approximately 0.11.

\vspace{-1.5ex}
\subsection{Frequency-Based Consensus Construction}
\vspace{-1ex}

As a secondary aggregation experiment, a frequency consensus $\mathcal{F}(\cdot)$ was constructed from the four most stable methods: JMIM, JMIM+CSNR, JMIM+CSNR+Corr, and Sim-LP. For each band $b$, its $\mathcal{F}(b)\in\{0,\ldots,40\}$ was computed across the top-$L$ subsets obtained from all ten ROI sets as:
\vspace{-1ex}
$$
\mathcal{F}(b)=\sum_{m\in\mathcal{M}_s}\sum_{r=1}^{10}
\mathcal{I}\left(b\in S_{m,r}^{L}\right),
\vspace{-1ex}
$$
and the corresponding subsets: $S_{\mathrm{Freq}}^K\in\underset{\substack{S\subseteq\mathcal{B}\\|S|=K}}{\arg\max}\sum_{b\in S}\mathcal{F}(b)$. Where $\mathcal{M}_s$ denotes the four retained methods, $\mathcal{B}$ is the set of all spectral bands, and $\mathcal{I}(\cdot)$ is the indicator function. Bands were then ranked in descending order of $\mathcal{F}(b)$, and the first $K$ bands formed $S_{\mathrm{Freq}}^K$, for $K\in\{3,5,\ldots,13\}$. These subsets were subsequently used for SSM training and evaluation.

\begin{figure*}[htp]
    \vspace{-1.5ex}
    \centering
        \includegraphics[
            width=\textwidth,
            trim={0 0 0 0},
            clip
        ]{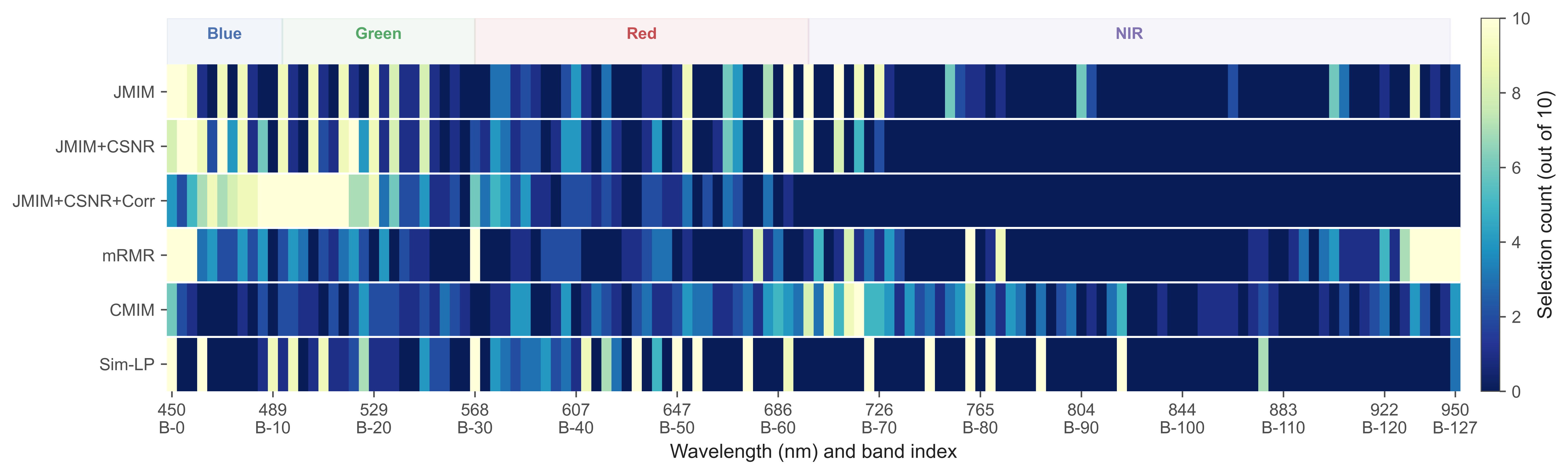} 
    \captionsetup{skip=2pt}

\caption{Method-specific top-25 band recurrence across ten class-balanced ROI sets. Color indicates selection count.}

    \label{fig:band_recurrence}
    \vspace{-3ex}
\end{figure*}

\vspace{-1.5ex}
\subsection{Semantic-Segmentation Validation}
\vspace{-1ex}
\label{sec:segmentation_validation}

For each method, rankings from the three ROI sets were truncated at $K$-bands for SSM evaluation. These sets were not selected based on segmentation performance and were limited to three for computational tractability, as using all ten would require 3.33$\times$ as many training runs. The subsets were evaluated with U-Net, DeepLabV3+, and SegFormer, representing convolutional encoder--decoder, atrous-convolutional, and transformer-based models, respectively. Each SSM--subset combination was trained with three training seeds. Thus, each SSM--method--$K$ configuration comprised nine evaluations (three ROI sets $\times$ three training seeds). ROI sampling determined the band rankings, whereas the training seeds captured segmentation training variation.

The frequency consensus produced one ranking from all ten ROI sets, whose top-$K$ subsets were evaluated with the same SSMs and three training seeds. The full 128-band input served as the SSM-specific baseline. SSM performance was measured using mean intersection over union (mIoU) and mean F1 score (mF1). Individual-method results are reported as mean $\pm$ standard deviation (SD) over nine evaluations, incorporating both ROI-set and SSM-training variation. Frequency-consensus and 128-band results are reported over three training seeds and therefore reflect only training variation. All configurations used the same preprocessing, training, and evaluation pipeline~\footnote{\url{https://github.com/imadalishah/multiseedHSI}}.

\vspace{-1.5ex}
\subsection{Method-Level Stability: Performance Comparison}
\vspace{-1ex}

For each method and subset size, the mean mIoU and mF1 obtained with each SSM were compared with the corresponding all-128-bands baseline: selected-band minus baseline, in percentage points. These differences were averaged across the three SSMs and six subset sizes for each method and metric, then compared with the method's mean over 45 pairwise Jaccard similarities. The frequency consensus was excluded because its single aggregated ranking has no intra-method stability value.

\begin{figure}[!htbp]
    \centering

        \includegraphics[
            width=0.5\textwidth,
            height=0.23\textheight,
            trim={0 0 0 0},
            clip
        ]{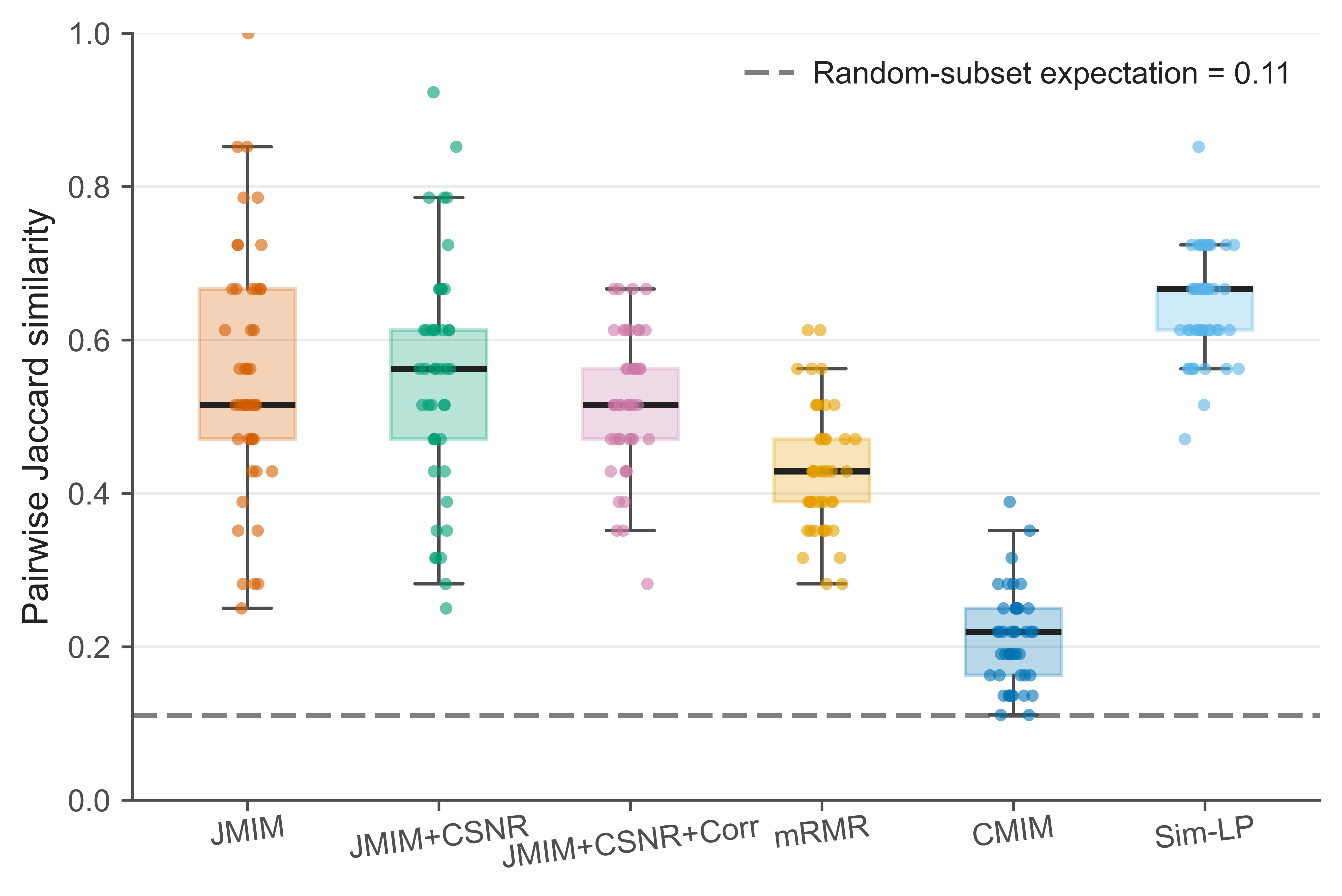}
    \captionsetup{skip=2pt}
    \caption{Intra-method pairwise Jaccard similarities of top-25 subsets across ten ROI sets. The dashed line marks the random-subset expectation ($0.11$).}
    \label{fig:jaccard_stability}
    \vspace{-2.25ex}
\end{figure}

\vspace{-1.5ex}
\section{Results and Discussion}
\vspace{-1ex}

\vspace{-1ex}
\subsection{Intra-Method Band-Selection Stability}
\vspace{-1ex}

\label{sec:stability_results}

Figure~\ref{fig:band_recurrence} shows that the recurrence patterns of the selected bands across the spectrum differ between methods. Sim-LP selects a sparse set distributed across the visible and near-infrared (NIR) ranges, whereas CMIM's recurrent selections span much of the spectrum, with several of its highest selection counts in the NIR. mRMR concentrates recurrence near both spectral boundaries. Compared with JMIM, JMIM+CSNR shifts recurrent selections toward the visible range, while JMIM+CSNR+Corr concentrates them in the blue-green region with little NIR recurrence.

Figure~\ref{fig:jaccard_stability} illustrates that the degree of agreement among the ten top-25 subsets also differs between methods. Sim-LP has the highest pairwise Jaccard similarity median and a relatively narrow interquartile range. JMIM, JMIM+CSNR, and JMIM+CSNR+Corr form a relatively high-agreement group below Sim-LP, although their distributions differ. mRMR has a lower median than this group, but remains above CMIM, which has the lowest median. The lowest CMIM values approach the random-subset reference of $0.11$. 

Overall, these results indicate that changing the ROI set affects the selected bands of each method differently, raising the question of whether methods with more consistent selections also achieve better segmentation performance.

\begin{table*}[!t]
\vspace{-1ex}
\centering
\caption{Mean $\pm$ SD mIoU and mF1 (\%) by SSM, method, and subset size. Individual methods use nine evaluations, while frequency-consensus and baseline entries use three. Cell color encodes the signed difference from SSM-specific 128-band baseline (green: positive; red: negative; white: parity), and bold marks each SSM-metric maximum.}
\vspace{-1.5ex}
\label{tab:inverted_performance}
\resizebox{\textwidth}{!}{%
\tiny
\setlength{\tabcolsep}{2pt}
\begin{tabular}{p{0.15cm}p{0.2cm} *{2}r | *{2}r | *{2}r | *{2}r | *{2}r | *{2}r | *{2}r | p{0.15cm}p{0.2cm}}
\hline
\multirow{2}{*}{\rotatebox{35}{\textbf{Model}}} &
\multirow{2}{*}{\rotatebox{35}{\textbf{Set Size}}} &
\multicolumn{2}{c}{\textbf{CMIM}} &
\multicolumn{2}{|c}{\textbf{JMIM}} &
\multicolumn{2}{|c}{\textbf{JMIM+CSNR}} &
\multicolumn{2}{|c}{\textbf{JMIM+CSNR+Corr}} &
\multicolumn{2}{|c}{\textbf{mRMR}} &
\multicolumn{2}{|c}{\textbf{Sim-LP}} &
\multicolumn{2}{|c}{\textbf{Freq.}} &
\multicolumn{2}{|c}{\textbf{128-band}} \\
& & \textbf{mIoU} & \textbf{mF1} & \textbf{mIoU} & \textbf{mF1} &
\textbf{mIoU} & \textbf{mF1} & \textbf{mIoU} & \textbf{mF1} &
\textbf{mIoU} & \textbf{mF1} & \textbf{mIoU} & \textbf{mF1} &
\textbf{mIoU} & \textbf{mF1} &
\rotatebox{25}{\scalebox{0.8}{\tiny\textbf{mIoU}}} &
\rotatebox{25}{\scalebox{0.8}{\tiny\textbf{mF1}}}\\
\hline
\multirow{6}{*}{\rotatebox{90}{\textbf{U-Net}}} & 3 & \heatcell{0.06}{70.73$\spm$1.23} & \heatcell{0.12}{79.68$\spm$1.32} & \heatcell{-2.24}{68.43$\spm$0.32} & \heatcell{-1.61}{77.95$\spm$0.25} & \heatcell{-1.88}{68.79$\spm$0.39} & \heatcell{-1.55}{78.01$\spm$0.31} & \heatcell{-2.91}{67.76$\spm$1.75} & \heatcell{-1.84}{77.72$\spm$1.23} & \heatcell{-2.05}{68.62$\spm$0.48} & \heatcell{-1.93}{77.63$\spm$0.69} & \heatcell{-0.79}{69.88$\spm$1.01} & \heatcell{-0.25}{79.31$\spm$0.73} & \heatcell{-1.01}{69.66$\spm$1.27} & \heatcell{-0.62}{78.94$\spm$0.88} & \multirow{6}{*}{\rotatebox{90}{70.67$\spm$0.61}} & \multirow{6}{*}{\rotatebox{90}{79.56$\spm$0.47}} \\
 & 5 & \heatcell{-0.61}{70.06$\spm$0.61} & \heatcell{-0.22}{79.34$\spm$0.43} & \heatcell{-0.73}{69.94$\spm$0.61} & \heatcell{-0.62}{78.94$\spm$0.73} & \heatcell{-2.16}{68.51$\spm$0.49} & \heatcell{-1.58}{77.98$\spm$0.23} & \heatcell{-2.64}{68.03$\spm$1.83} & \heatcell{-1.82}{77.74$\spm$1.61} & \heatcell{0.12}{70.79$\spm$0.76} & \heatcell{0.29}{79.85$\spm$0.25} & \heatcell{-0.66}{70.01$\spm$0.77} & \heatcell{0.11}{79.67$\spm$0.60} & \heatcell{-1.70}{68.97$\spm$0.85} & \heatcell{-0.91}{78.65$\spm$0.77} &  &  \\
 & 7 & \heatcell{-0.06}{70.61$\spm$0.31} & \heatcell{0.34}{79.90$\spm$0.20} & \heatcell{-0.65}{70.02$\spm$0.23} & \heatcell{-0.31}{79.25$\spm$0.95} & \heatcell{-0.14}{70.53$\spm$1.19} & \heatcell{-0.04}{79.52$\spm$0.88} & \heatcell{-4.03}{66.64$\spm$0.85} & \heatcell{-3.00}{76.56$\spm$0.67} & \heatcell{-0.66}{70.01$\spm$0.29} & \heatcell{-0.06}{79.50$\spm$0.12} & \heatcell{-0.28}{70.39$\spm$1.36} & \heatcell{0.36}{79.92$\spm$1.20} & \heatcell{-0.70}{69.97$\spm$1.32} & \heatcell{-0.45}{79.11$\spm$1.42} &  &  \\
 & 9 & \heatcell{0.10}{70.77$\spm$0.37} & \heatcell{0.10}{79.66$\spm$0.50} & \heatcell{-0.33}{70.34$\spm$1.28} & \heatcell{0.07}{79.63$\spm$1.07} & \heatcell{0.43}{\textbf{71.10$\spm$1.09}} & \heatcell{0.54}{80.10$\spm$0.94} & \heatcell{-2.69}{67.98$\spm$0.98} & \heatcell{-1.72}{77.84$\spm$0.79} & \heatcell{0.24}{70.91$\spm$1.04} & \heatcell{0.02}{79.58$\spm$1.09} & \heatcell{0.38}{71.05$\spm$0.32} & \heatcell{0.70}{\textbf{80.26$\spm$0.50}} & \heatcell{-0.71}{69.96$\spm$0.17} & \heatcell{-0.55}{79.01$\spm$0.15} &  &  \\
 & 11 & \heatcell{-0.65}{70.02$\spm$1.75} & \heatcell{-0.22}{79.34$\spm$1.10} & \heatcell{-2.00}{68.67$\spm$1.23} & \heatcell{-1.23}{78.33$\spm$1.09} & \heatcell{-0.33}{70.34$\spm$0.43} & \heatcell{0.25}{79.81$\spm$0.55} & \heatcell{-2.52}{68.15$\spm$0.38} & \heatcell{-1.60}{77.96$\spm$0.72} & \heatcell{-0.78}{69.89$\spm$0.27} & \heatcell{-0.43}{79.13$\spm$0.52} & \heatcell{-1.58}{69.09$\spm$1.23} & \heatcell{-0.68}{78.88$\spm$1.05} & \heatcell{-1.85}{68.82$\spm$1.10} & \heatcell{-1.45}{78.11$\spm$0.92} &  &  \\
 & 13 & \heatcell{-0.39}{70.28$\spm$0.62} & \heatcell{-0.46}{79.10$\spm$0.48} & \heatcell{-1.01}{69.66$\spm$0.37} & \heatcell{-0.65}{78.91$\spm$0.09} & \heatcell{-0.85}{69.82$\spm$0.10} & \heatcell{-0.60}{78.96$\spm$0.36} & \heatcell{-3.67}{67.00$\spm$0.44} & \heatcell{-2.59}{76.97$\spm$0.41} & \heatcell{-0.20}{70.47$\spm$0.91} & \heatcell{0.14}{79.70$\spm$0.69} & \heatcell{-0.05}{70.62$\spm$0.73} & \heatcell{0.04}{79.60$\spm$0.24} & \heatcell{-2.07}{68.60$\spm$0.27} & \heatcell{-1.45}{78.11$\spm$0.17} &  &  \\
\hline
\multirow{6}{*}{\rotatebox{90}{\textbf{DLV3+}}} & 3 & \heatcell{0.82}{70.89$\spm$0.98} & \heatcell{0.56}{79.90$\spm$0.82} & \heatcell{-0.94}{69.13$\spm$0.87} & \heatcell{-0.96}{78.38$\spm$0.65} & \heatcell{-0.40}{69.67$\spm$0.66} & \heatcell{-0.29}{79.05$\spm$0.39} & \heatcell{-2.05}{68.02$\spm$1.07} & \heatcell{-1.80}{77.54$\spm$0.95} & \heatcell{-1.74}{68.33$\spm$0.42} & \heatcell{-1.53}{77.81$\spm$0.66} & \heatcell{1.02}{\textbf{71.09$\spm$0.67}} & \heatcell{1.05}{80.39$\spm$0.83} & \heatcell{-0.60}{69.47$\spm$0.66} & \heatcell{-0.49}{78.85$\spm$0.71} & \multirow{6}{*}{\rotatebox{90}{70.07$\spm$0.66}} & \multirow{6}{*}{\rotatebox{90}{79.34$\spm$0.24}} \\
 & 5 & \heatcell{0.34}{70.41$\spm$1.18} & \heatcell{0.43}{79.77$\spm$1.05} & \heatcell{0.46}{70.53$\spm$1.00} & \heatcell{0.17}{79.51$\spm$0.88} & \heatcell{-1.15}{68.92$\spm$0.30} & \heatcell{-0.90}{78.44$\spm$0.34} & \heatcell{-2.32}{67.75$\spm$1.04} & \heatcell{-2.08}{77.26$\spm$1.27} & \heatcell{0.07}{70.14$\spm$1.34} & \heatcell{0.33}{79.67$\spm$1.17} & \heatcell{-0.08}{69.99$\spm$0.45} & \heatcell{-0.15}{79.19$\spm$0.63} & \heatcell{-1.20}{68.87$\spm$0.08} & \heatcell{-1.27}{78.07$\spm$0.16} &  &  \\
 & 7 & \heatcell{-0.17}{69.90$\spm$0.72} & \heatcell{-0.20}{79.14$\spm$0.44} & \heatcell{0.08}{70.15$\spm$0.38} & \heatcell{0.09}{79.43$\spm$0.33} & \heatcell{-0.51}{69.56$\spm$0.44} & \heatcell{-0.09}{79.25$\spm$0.60} & \heatcell{-2.62}{67.45$\spm$0.33} & \heatcell{-2.04}{77.30$\spm$0.28} & \heatcell{-1.40}{68.67$\spm$1.02} & \heatcell{-1.03}{78.31$\spm$0.87} & \heatcell{0.07}{70.14$\spm$0.61} & \heatcell{-0.06}{79.28$\spm$0.36} & \heatcell{-1.15}{68.92$\spm$0.60} & \heatcell{-1.07}{78.27$\spm$0.26} &  &  \\
 & 9 & \heatcell{0.72}{70.79$\spm$0.46} & \heatcell{0.39}{79.73$\spm$0.55} & \heatcell{0.03}{70.10$\spm$0.61} & \heatcell{0.30}{79.64$\spm$0.43} & \heatcell{0.37}{70.44$\spm$0.54} & \heatcell{0.33}{79.67$\spm$0.39} & \heatcell{-2.74}{67.33$\spm$0.34} & \heatcell{-2.21}{77.13$\spm$0.77} & \heatcell{-0.37}{69.70$\spm$0.85} & \heatcell{-0.18}{79.16$\spm$0.89} & \heatcell{0.95}{71.02$\spm$0.59} & \heatcell{1.28}{\textbf{80.62$\spm$0.48}} & \heatcell{-0.83}{69.24$\spm$1.00} & \heatcell{-0.57}{78.77$\spm$1.06} &  &  \\
 & 11 & \heatcell{0.55}{70.62$\spm$0.72} & \heatcell{0.05}{79.39$\spm$1.09} & \heatcell{0.35}{70.42$\spm$0.85} & \heatcell{0.07}{79.41$\spm$0.55} & \heatcell{-0.19}{69.88$\spm$0.49} & \heatcell{0.02}{79.36$\spm$0.21} & \heatcell{-1.41}{68.66$\spm$0.30} & \heatcell{-1.17}{78.17$\spm$0.37} & \heatcell{0.20}{70.27$\spm$0.48} & \heatcell{0.19}{79.53$\spm$0.10} & \heatcell{0.09}{70.16$\spm$0.11} & \heatcell{0.18}{79.52$\spm$0.12} & \heatcell{-1.21}{68.86$\spm$0.80} & \heatcell{-1.33}{78.01$\spm$1.02} &  &  \\
 & 13 & \heatcell{0.06}{70.13$\spm$1.00} & \heatcell{0.05}{79.39$\spm$0.69} & \heatcell{-0.34}{69.73$\spm$0.61} & \heatcell{-0.36}{78.98$\spm$0.59} & \heatcell{-0.24}{69.83$\spm$1.64} & \heatcell{-0.08}{79.26$\spm$1.23} & \heatcell{-1.64}{68.43$\spm$0.52} & \heatcell{-1.37}{77.97$\spm$0.50} & \heatcell{-0.72}{69.35$\spm$0.66} & \heatcell{-0.66}{78.68$\spm$0.62} & \heatcell{0.80}{70.87$\spm$1.02} & \heatcell{0.83}{80.17$\spm$1.01} & \heatcell{-1.51}{68.56$\spm$0.30} & \heatcell{-1.40}{77.94$\spm$0.41} &  &  \\
\hline
\multirow{6}{*}{\rotatebox{90}{\textbf{SegFormer}}} & 3 & \heatcell{0.49}{67.99$\spm$0.81} & \heatcell{0.44}{77.50$\spm$0.50} & \heatcell{-1.66}{65.84$\spm$0.93} & \heatcell{-1.20}{75.86$\spm$1.02} & \heatcell{-0.42}{67.08$\spm$0.87} & \heatcell{-0.47}{76.59$\spm$0.37} & \heatcell{-2.40}{65.10$\spm$0.81} & \heatcell{-1.73}{75.33$\spm$1.04} & \heatcell{-1.86}{65.64$\spm$1.34} & \heatcell{-1.63}{75.43$\spm$0.56} & \heatcell{1.26}{68.76$\spm$0.79} & \heatcell{1.15}{78.21$\spm$0.92} & \heatcell{-1.26}{66.24$\spm$2.03} & \heatcell{-1.15}{75.91$\spm$1.78} & \multirow{6}{*}{\rotatebox{90}{67.50$\spm$1.05}} & \multirow{6}{*}{\rotatebox{90}{77.06$\spm$0.60}} \\
 & 5 & \heatcell{1.66}{69.16$\spm$0.84} & \heatcell{1.33}{78.39$\spm$0.62} & \heatcell{0.97}{68.47$\spm$1.10} & \heatcell{1.07}{78.13$\spm$0.72} & \heatcell{1.25}{68.75$\spm$0.49} & \heatcell{1.48}{78.54$\spm$0.54} & \heatcell{-2.16}{65.34$\spm$2.10} & \heatcell{-1.75}{75.31$\spm$1.56} & \heatcell{-0.36}{67.14$\spm$0.29} & \heatcell{0.32}{77.38$\spm$0.18} & \heatcell{1.47}{68.97$\spm$0.47} & \heatcell{1.72}{\textbf{78.78$\spm$0.52}} & \heatcell{-0.50}{67.00$\spm$0.38} & \heatcell{-0.60}{76.46$\spm$0.22} &  &  \\
 & 7 & \heatcell{0.18}{67.68$\spm$0.40} & \heatcell{0.58}{77.64$\spm$0.44} & \heatcell{1.07}{68.57$\spm$0.25} & \heatcell{0.86}{77.92$\spm$0.17} & \heatcell{2.01}{\textbf{69.51$\spm$0.27}} & \heatcell{1.59}{78.65$\spm$0.62} & \heatcell{-1.29}{66.21$\spm$0.55} & \heatcell{-0.71}{76.35$\spm$0.71} & \heatcell{-0.42}{67.08$\spm$0.46} & \heatcell{-0.52}{76.54$\spm$0.25} & \heatcell{0.30}{67.80$\spm$1.27} & \heatcell{0.55}{77.61$\spm$0.66} & \heatcell{0.05}{67.55$\spm$0.31} & \heatcell{-0.14}{76.92$\spm$0.23} &  &  \\
 & 9 & \heatcell{1.47}{68.97$\spm$0.38} & \heatcell{1.55}{78.61$\spm$0.55} & \heatcell{1.06}{68.56$\spm$0.36} & \heatcell{0.85}{77.91$\spm$0.18} & \heatcell{1.11}{68.61$\spm$0.86} & \heatcell{1.26}{78.32$\spm$0.68} & \heatcell{-2.44}{65.06$\spm$0.91} & \heatcell{-1.96}{75.10$\spm$0.64} & \heatcell{-0.62}{66.88$\spm$2.04} & \heatcell{-0.51}{76.55$\spm$1.25} & \heatcell{0.74}{68.24$\spm$0.67} & \heatcell{0.80}{77.86$\spm$0.19} & \heatcell{-1.43}{66.07$\spm$0.85} & \heatcell{-1.08}{75.98$\spm$0.82} &  &  \\
 & 11 & \heatcell{-0.28}{67.22$\spm$1.63} & \heatcell{0.43}{77.49$\spm$1.34} & \heatcell{0.61}{68.11$\spm$1.39} & \heatcell{0.66}{77.72$\spm$0.88} & \heatcell{1.19}{68.69$\spm$1.17} & \heatcell{1.01}{78.07$\spm$1.24} & \heatcell{-1.14}{66.36$\spm$1.62} & \heatcell{-0.76}{76.30$\spm$1.05} & \heatcell{0.18}{67.68$\spm$1.88} & \heatcell{0.41}{77.47$\spm$1.85} & \heatcell{1.41}{68.91$\spm$0.61} & \heatcell{1.46}{78.52$\spm$0.62} & \heatcell{-0.88}{66.62$\spm$0.18} & \heatcell{-1.28}{75.78$\spm$0.40} &  &  \\
 & 13 & \heatcell{0.11}{67.61$\spm$0.50} & \heatcell{0.27}{77.33$\spm$0.44} & \heatcell{0.89}{68.39$\spm$0.87} & \heatcell{0.98}{78.04$\spm$0.97} & \heatcell{1.18}{68.68$\spm$0.81} & \heatcell{1.14}{78.20$\spm$0.79} & \heatcell{-1.93}{65.57$\spm$1.73} & \heatcell{-1.57}{75.49$\spm$1.51} & \heatcell{-0.69}{66.81$\spm$0.32} & \heatcell{-0.30}{76.76$\spm$0.71} & \heatcell{1.12}{68.62$\spm$0.48} & \heatcell{0.87}{77.93$\spm$0.64} & \heatcell{-0.01}{67.49$\spm$0.89} & \heatcell{-0.39}{76.67$\spm$0.96} &  &  \\
\hline
\end{tabular}%
}
\vspace{-0.5ex}
\end{table*}

\vspace{-1.5ex}
\subsection{Segmentation Performance of Band Subsets}
\vspace{-1ex}

Table~\ref{tab:inverted_performance} shows that Sim-LP and JMIM+CSNR achieve all six SSM-metric best results. For U-Net, JMIM+CSNR and Sim-LP at $K=9$ improve mIoU and mF1 by 0.43 and 0.70 percentage points over the baseline, respectively. Sim-LP gives the best results for DeepLabV3+, improving mIoU by 1.02 points at $K=3$ and mF1 by 1.28 points at $K=9$. The largest gains occur with SegFormer: 2.01 mIoU points for JMIM+CSNR at $K=7$ and 1.72 mF1 points for Sim-LP at $K=5$.

\begin{figure}[!htbp]
    \centering
    \includegraphics[width=\linewidth,]{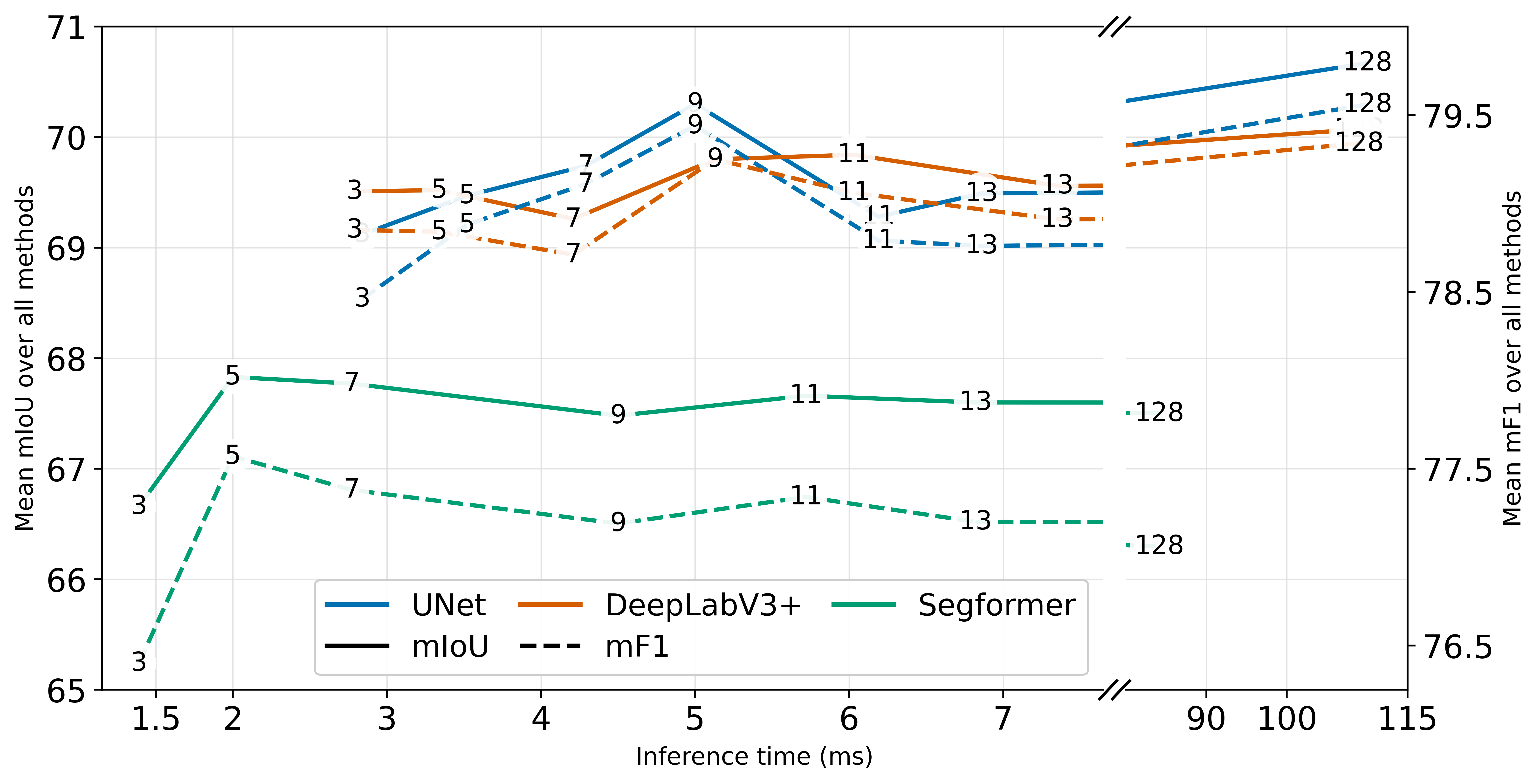}
    \vspace{-4ex}
    \caption{CPU input-layer inference time and averaged mIoU/mF1 across methods in Table~\ref{tab:inverted_performance}: Annotations indicate input and 128-band points. Inference used batch size 1, 10 warm-up iterations, excluded data loading, and was benchmarked on Intel Xeon Gold 6252, Debian 12.}
    \label{fig:inferenceperformance}
    \vspace{-1ex}
\end{figure}
\begin{figure}[!h]
    \centering
    \includegraphics[width=\linewidth,]{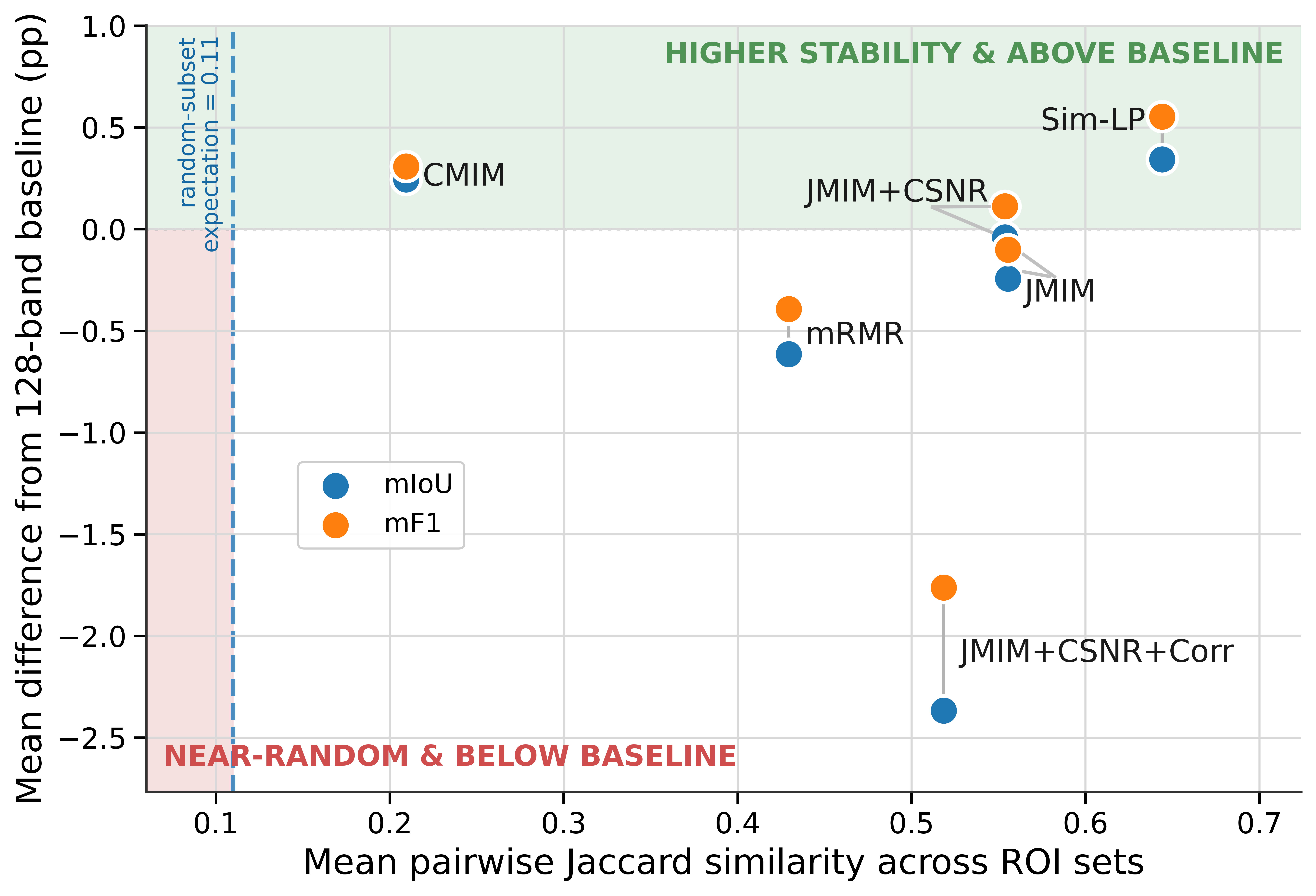}
    \vspace{-3ex}
    \caption{Mean pairwise Jaccard similarity versus baseline-relative mIoU/mF1 differences, averaged across SSMs and subset sizes. Differences are reported in percentage points. Dashed lines mark the random-subset expectation ($0.11$) and baseline parity. Shading indicates positive performance differences and stability below the random-subset reference.}
    \label{fig:stability_utility}
    \vspace{-2ex}
\end{figure}

$K=9$ yields three of the six best results and the largest number of above-baseline results (25 of 36). However, performance is non-monotonic with $K$, and no subset size is optimal across all SSMs and metrics. The frequency consensus exceeds baseline in only one of 36 results and never surpasses the best individual method, indicating that selection frequency of stable methods alone does not identify the highest-performing subsets.

\subsection{Inference-Time and Accuracy Trade-off}
Figure~\ref{fig:inferenceperformance} shows that spectral reduction substantially reduces CPU input-layer inference time while largely preserving segmentation performance. At $K=9$, inference is approximately $18.7$--$22.0\times$ faster than with 128 bands across the three SSMs. Averaged across the six band-selection methods, mean performance at $K=9$ remains close to baseline (mIoU and mF1 differences within roughly $\pm$0.1 percentage points). However, this aggregate masks substantial method-level variation: five of six methods stay within about 1 percentage point of baseline, while JMIM+CSNR+Corr trails baseline by 2.44--2.74 mIoU and 1.72--2.21 mF1 points. These results indicate that $K=9$ provides a favorable aggregate accuracy--efficiency trade-off, although the optimal subset size remains method- and SSM-dependent.

\vspace{-1.5ex}
\subsection{Method-Level Stability--Performance Association}
\vspace{-1ex}

Figure~\ref{fig:stability_utility} shows no monotonic relationship between intra-method stability and baseline-relative segmentation performance. Sim-LP combines the highest stability with the largest positive method-level differences: approximately 0.34 percentage points in mIoU and 0.55 percentage points in mF1. JMIM+CSNR remains close to the corresponding baseline (-0.04 mIoU and +0.11 mF1), and JMIM+CSNR+Corr has the largest performance deficits (-2.37 mIoU and -1.76 mF1) despite relatively high stability. Whereas CMIM has the lowest stability but positive differences in both metrics. JMIM+CSNR remains close to baseline despite relatively high stability. The results indicate that greater stability neither guarantees better segmentation performance nor is required for it, and the two criteria should be assessed jointly for band selection performance studies.

\vspace{-1.5ex}
\section{Conclusion}
\label{sec:conclusion}
\vspace{-1ex}

This study evaluated the intra-method stability of six band-selection methods under repeated class-balanced ROI sampling. It examined whether stability was associated with downstream semantic-segmentation performance across subset sizes and SSMs. Repeated class-balanced ROI sampling showed that band-selection stability is method-dependent but does not reliably predict downstream segmentation performance. Sim-LP remained the most intra-method stable, while Sim-LP and JMIM+CSNR produced the six SSM-metric optimal results, including gains of up to 2.01 mIoU and 1.72 mF1 points over 128-band baselines. The relatively stable JMIM+CSNR+Corr showed the largest performance deficits, while the least stable CMIM method showed positive average baseline-relative differences in both metrics. Three of the best results occurred at $K=9$, while the others occurred once each in $K\in\{3,5,7\}$. The frequency-consensus ranking derived from the more stable methods did not outperform individual methods, indicating that stability-driven consensus alone is insufficient for selecting high-performing subsets. Performance was non-monotonic with subset size, although $K=9$ provided a favorable aggregate efficiency–accuracy trade-off, with 18.7–22.0$\times$ faster CPU input-layer inference while remaining close to baseline performance.

These results show that intra-method stability is useful for characterizing the reproducibility of a band-selection method, but is not a reliable indicator of downstream segmentation performance in these experiments. Band-selection methods should therefore be evaluated across repeated samples, subset sizes, and downstream SSMs rather than judged by stability or a single selected subset alone. These conclusions are limited to one urban HSI dataset and pairwise Jaccard similarity. Future work should assess whether the observed stability--performance relationship generalizes across additional HSI segmentation datasets and alternative stability measures.

\vspace{-2ex}
\let\oldthebibliography\thebibliography
\let\endoldthebibliography\endthebibliography
\renewenvironment{thebibliography}[1]{%
  \begin{oldthebibliography}{#1}%
    \setlength{\itemsep}{0pt}
    \setlength{\parsep}{0pt}
    \setlength{\baselineskip}{11.6pt}
}{%
  \end{oldthebibliography}%
}

\bibliographystyle{IEEEtran}
\bibliography{strings,refs}
\end{document}